\documentclass[10pt,twocolumn,letterpaper]{article}

\usepackage{cvpr} 
\usepackage{times}
\usepackage{epsfig}
\usepackage{graphicx}
\usepackage{amsmath}
\usepackage{amssymb}

\begin{document}
\twocolumn[{%
\title{Gradient-Guided Exploration of Decoder-Based Generative Model's Latent Space for Controlled Iris Image Augmentations}

\author{Mahsa Mitcheff, Siamul Karim Khan, and Adam Czajka\\
384 Fitzpatrick Hall of Engineering, University of Notre Dame, IN 46556, USA\\
{\tt\small \{mmitchef,skhan22,aczajka\}@nd.edu}}

\maketitle
\thispagestyle{empty}

\begin{center}
    \includegraphics[width=\textwidth]{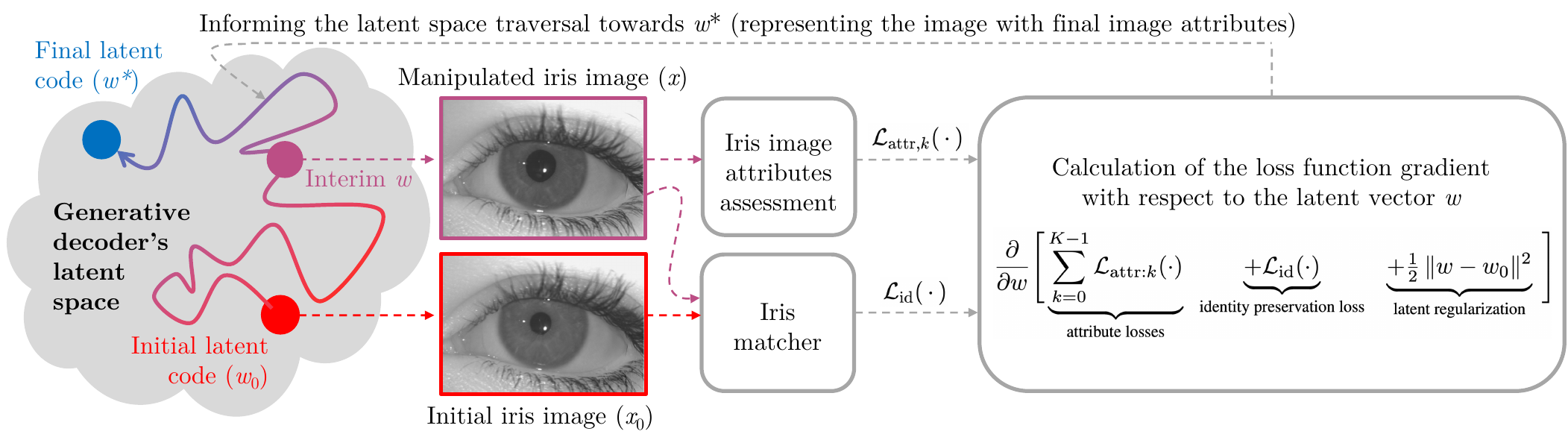}
    
    \vspace{2mm}
    \captionof{figure}{We manipulate selected geometrical or textural iris image attribute by traversing the latent space of a generative decoder (trained to synthesize ISO/IEC 19794-6-compliant iris images). This traversal is guided by the gradient of a multi-term loss function (including an identity preservation component) with respect to the decoder's latent space.}
    \label{fig:teaser}
\end{center}
\vspace{5mm}
}]

\begin{abstract}
   Developing reliable iris recognition and presentation attack detection methods requires diverse datasets that capture realistic variations in iris features and a wide spectrum of anomalies. Because of the rich texture of iris images, which spans a wide range of spatial frequencies, synthesizing same-identity iris images while controlling specific attributes remains challenging. In this work, we introduce a new iris image augmentation strategy by traversing the latent space of a pre-trained decoder-based generative model toward latent codes that represent same-identity samples but with selected iris image properties manipulated. The latent space traversal is (a) guided by a gradient of specific geometrical, textural, or quality-related iris image features (\eg, sharpness, pupil size, iris size, or pupil-to-iris ratio) with respect to the latent codes, and (b) preserves the identity represented by the image being manipulated. The proposed approach can be easily extended to manipulate any attribute for which a differentiable loss term can be formulated. The source codes are offered with the paper. 
\end{abstract}

\section{Introduction}
\label{sec:intro}

\subsection{Background and Motivation}
Iris recognition is widely recognized as a highly accurate, reliable and mature biometric modality, with the first working algorithm developed in 1993 \cite{daugman1993high}. While (owing to a mathematically-elegant and grounded in information theory and neuroscience) iris feature encoding proposed by Daugman did not observe vast changes over decades, the other elements of iris recognition pipeline, such as image segmentation or presentation attack detection, benefit from deep learning, which may generalize better than hand-crafted algorithms to unseen anomalies. However, the development of robust deep learning-based models is hindered by the significant costs and privacy issues associated with collecting large-scale, diverse iris datasets \cite{wang2022generating}. To mitigate these challenges, researchers have increasingly utilized synthetically generated iris images, employing both traditional methods and modern Generative Artificial Intelligence (GenAI) models.

One of the still-unsolved challenges in iris image synthesis is the difficulty in generating same-eye samples with selected geometrical or textural image features controlled (\eg, pupil dilation, eyelid coverage, or off-axis gaze simulation). Such capability is essential for making iris image synthesis useful in advanced data augmentations to train more robust iris recognition and iris presentation attack detection methods that can be exposed to real-world, yet unseen image variations. In the past, synthetic image-based iris data augmentation relied on parametric image augmentation models, altering simple geometric image features \cite{wei2008synthesis}. State-of-the-art methods use non-parametric models (mostly deep learning-based) \cite{yadav2023iwarpgan, wang2022generating, khan2023deformirisnet}, where simultaneously controlling specific image properties while preserving the identity represented in an altered iris image remains a significant challenge.

\subsection{Proposed Solution}
We benefit from deep decoders, such as those used in StyleGAN architectures, capable of synthesizing high-fidelity iris images, and we propose a method that traverses the decoder's latent space towards a point corresponding to a synthesized image with the desired properties, as illustrated in Fig. \ref{fig:teaser}. This traversal is guided by the gradient of differentiable functions that assess a set of iris images features, which gives a precise control over individual geometrical, textural and identity-related iris attributes. If the desired image properties are non-differentiable with respect to the latent space, they may be replaced with differentiable proxies or pretrained differentiable regressors, which makes this method quite generic and applicable to a wide family of visual data augmentations, going beyond iris recognition, or biometrics in general.

An interesting advantage of the proposed gradient-guided latent space exploration is that it can be applied to any pre-trained generative model utilizing a decoder-based generator mapping a latent code to an image, if the gradient of the function measuring the desired image property exists. This includes generative models, which do not offer strong feature representation disentanglement in the latent space.

\subsection{Summary of Contributions}
\noindent
Summarizing, this work offers two key contributions:

\begin{enumerate}
    \item[a)] an approach for identity-preserving iris image synthesis with image attribute manipulation controlled by gradient-guided traversal of a latent space of a decoder-based generative model (the novelty of this approach compared to previous works in explained in Sec. \ref{related:novelty}),
    \item[b)] implementation\footnote{\url{httsp://github.com/...} GitHub address has been removed to preserve anonymity, and will be added in case this paper is accepted} that generates and manipulates full-resolution ($640\times480$ pixels) iris images compliant with ISO/IEC 19794-6, which are directly suitable for use in iris recognition pipelines and standardized evaluation protocols.
\end{enumerate}

The proposed method is applicable to any pre-trained generator of iris images utilizing decoder network mapping latent codes into iris images. The method is also self-supervised and does not require labeled data to enable attribute manipulation.
\section{Prior and Related Works}

\subsection{Non-Deep Learning-Based Iris Synthesis} 

These approaches can be broadly classified into two categories: methods that generate new iris images and those that synthesize images by assembling patches from the existing iris textures. From-scratch techniques utilize Principal Component Analysis \cite{cui2004iris}, Markov Random Fields \cite{makthal2005synthesis, shah2006generating}, anatomical modeling \cite{zuo2006model, zuo2007generation}, and multi-resolution techniques such as reverse subdivision \cite{wecker2005iris, wecker2010multiresolution}. Patch-based methods construct synthetic irises by combining segments from authentic iris textures \cite{wei2008synthesis}. A significant limitation across these non-deep learning methods is the unrealistic appearance of the synthesized irises, relatively high computational complexity of the algorithms, and modeling of iris image properties that are interpretable and understood the algorithm designers \cite{wei2008synthesis}. 

\subsection{Deep Learning-Based Iris Synthesis} 

To overcome limitations of non-deep learning approaches, researchers have increasingly utilized modern generative models, which are not restricted to any particular anatomical model of an eye and can learn the correct appearance of the frontal eye surface directly from the data. Current implementations of generative AI models synthesize iris images either starting from a random vector in model's latent space which is synthesizing an iris image representing unknown identity, or by transferring the style of one iris to another existing iris image.

A wide array of generative adversarial network (GAN) architectures and their variants have been proposed in the first group, including deep convolutional GAN \cite{minaee2018iris,kohli2017synthetic}, relativistic average standard GAN \cite{yadav2019synthesizing}, as well as more advanced models like conditional GANs \cite{lee2019conditional}, Wasserstein GAN with gradient penalty \cite{gulrajani2017improved}, and conditional Wasserstein GAN with gradient penalty \cite{li2021conditional}. Cyclic image translation GAN \cite{yadav2021cit}, DeformIrisNet \cite{khan2023deformirisnet}, and iWarpGAN \cite{yadav2023iwarpgan} are examples of style-transfer methods. 

Unlike traditional techniques, generative AI models can produce photorealistic images that accurately reflect the distribution of the training data. However, these models come with significant drawbacks, primarily the high computational expense of training and the necessity of very large training datasets. For a detailed review of both traditional and generative AI methods for iris synthesis, we recommend two surveys by Yadav \etal \cite{yadav2024synthesizing} and Sawilska \etal \cite{sawilska2025synthetic}.

\subsection{Mitigating the Intra-class Diversity Challenge in Iris Synthesis}
\label{mitigation diversity}

One of the primary challenges with using synthetic iris images generated by GenAI is their limited capacity to produce diverse intra-class (\ie same-eye) variations. There are two main approaches for addressing this: transformation techniques, which apply image augmentations to a given iris image to create new variations, and style transformation, which transfers texture or structural patterns from one iris to another to generate a new sample. 

Wei \etal \cite{wei2008synthesis} introduced a series of transformations to enhance the intra-class variation of synthetic iris images. Their method employs non-linear deformation to create irises with varying pupil sizes, which simulates the texture distortions caused by pupil dilation and contraction. To further replicate real-world imaging conditions, additional transformations are applied. These include Gaussian blurring to simulate defocus, random pixel perturbation with bilinear interpolation, and iris rotation achieved through horizontal translation in a normalized polar coordinate system \cite{wei2008synthesis}. 

Yadav \etal \cite{yadav2021cit} introduced the CIT-GAN, a novel multi-domain style transfer GAN designed to generate high-quality synthetic images for iris presentation attack detection. CIT-GAN enhances the standard GAN by improving its ability to translate bonafide iris images across different presentation attack categories (\eg contact lens, printed and doll eye). It has a styling network that captures domain-specific traits, enabling the generator to modify existing images rather than creating them from random noise. Additionally, it includes a multi-branch discriminator for domain-aware classification and cycle consistency loss to ensure transformed images retain key features, making it highly effective for multi-domain image translation in iris presentation attack detection. 

Wang \etal \cite{wang2022generating} incorporated contrastive learning techniques to effectively disentangle identity-related features from condition-variant features, such as pupil size. While this approach successfully produces iris images with diverse intra- and inter-class variations, the study's focus is limited to feature disentanglement. 

DeformIrisNet introduced by Khan \etal \cite{khan2023deformirisnet} is a deep autoencoder based on the U-Net architecture for modifying pupil size in near-infrared, ISO-compliant iris images. The model takes a source iris image and a binary mask defining the target iris shape as input, enabling it to apply nonlinear texture deformations that simulate variations in pupil size. A limitation of this approach is its inability to preserve the identity of the original iris, as the deformations can alter key biometric features.  

Yadav \etal \cite{yadav2023iwarpgan} proposed iWarpGAN, a model designed to disentangle identity and style for generating cropped iris images with both existing and non-existing identities. Specifically, iWarpGAN is capable of two key operations: generating new samples of a given identity by adopting the style of a reference image, and creating images with a different identity while preserving the style of the input iris image. Note that iWarpGAN requires a reference image (in addition to an image being altered) to transfer the style from.

\subsection{Gradient-Guided Latent Space Exploration}

Several approaches have been proposed to traverse the latent spaces of GAN models trained on face, animals, and animated images, enabling controlled manipulation of the image attributes \cite{shen2020interpreting, natsume2018rsgan}. Tzelepis \etal \cite{tzelepis2021warpedganspace} proposed an unsupervised approach that discovers non-linear, interpretable paths in the latent space of pretrained GANs. Their method employs Radial Basic Function-based warping functions to learn a path in the latent space, which outperform linear paths by producing more disentangled and semantically-meaningful transformations (\eg pose, expressions). More recently, Song \etal~\cite{song2023latent} introduced a physics-inspired framework that models latent traversals as gradient flows within dynamic potential landscapes. By learning multiple distinct and semantically consistent potentials, their method achieves disentangled and flexible trajectories. Moreover, it can be integrated as a regularizer during training, thereby encouraging structured latent representations and improving likelihood in GANs and VAEs. Pernus \etal~\cite{pernuvs2023maskfacegan} introduced MaskFaceGAN, a GAN-inversion method for local facial attribute editing that travels in the StyleGAN2 latent space using face-parser masks for spatial control and an attribute classifier for target semantics. In the end, the edited result will blend with the original image to add the background and other facial components not considered during optimization.

However, there is a research gap in applying gradient-guided traversal of a latent space of a pre-trained decoder-based generative model for altering iris image properties. Manipulation of iris attributes poses unique challenges due to the richer and more subtle set of discernible features compared to those of faces. For example, for face editing with modern generative models, it is possible to identify a hyperplane in the latent space that acts as a decision boundary separating selected semantic facial attributes~\cite{shen2020interfacegan, tzelepis2021warpedganspace}, which was not yet demonstrated in iris image synthesis. 

\subsection{How This Paper Differs from Previous Works}
\label{related:novelty}

Unlike existing style-transfer methods such as iWarpGAN \cite{yadav2023iwarpgan} and DeformIrisNet \cite{khan2023deformirisnet}, which require an external reference image or a binary mask to define the target iris shape, and approaches like CIT-GAN \cite{yadav2021cit} and Wang \etal \cite{wang2022generating}, which depend on training data consisting of irises with different identities or styles, our proposed method is entirely reference-free. It directly modifies the iris attributes by operating on the latent code of a given image. Also, most existing methods—except DeformIrisNet \cite{khan2023deformirisnet}—do not offer explicit control over the degree of variation in iris features. In contrast, our approach enables adjustment of feature size to a specified target value.
 
In addition, the models reviewed in Sec.~\ref{mitigation diversity} generate center-cropped iris images that are not compliant with ISO/IEC 19794-6 standards. Consequently, they cannot be integrated into iris biometric pipelines.

\section{Solution Description}
\subsection{Problem Formulation}

Let $G : \mathcal{W} \to \mathcal{X}$ denote a differentiable generative decoder 
that maps latent codes to iris images, where $\mathcal{W} \subset \mathbb{R}^d$ is the $d$-dimensional latent space, and $\mathcal{X} \subset \mathbb{R}^{H \times W}$ is the image space. $W$ and $H$ are iris image width and height, which for ISO/IEC 19794-6-compliant iris samples are 640 and 480 pixels, respectively.

We randomly pick an initial latent code $w_0 \in \mathcal{W}$, which corresponds to a synthetic image $x_0 = G(w_0)$. Let's consider a collection of attributes $\{ \mathcal{A}_k \}_{k=1}^K$ that may be manipulated (\eg, iris image sharpness, pupil radius, iris radius, or pupil-to-iris ratio). Each attribute $\mathcal{A}_k$ is associated with a differentiable scalar 
function:
\begin{displaymath}
    a_k : \mathcal{X} \to \mathbb{R},
\end{displaymath}
which measures the value of the attribute for a generated image.

To preserve the biometric identity during attribute manipulation, we employ a differentiable iris identity encoding function:
\begin{displaymath}
    \phi_{\mathrm{id}} : \mathcal{X} \to \mathbb{R}^m, 
\end{displaymath}

\noindent
where $m$ is the dimensionality of the iris template vector. 

\subsection{Composite Loss Formulation}

Our goal is to find a latent representation $w^\star$ such that the synthesized image $x^\star = G(w^\star)$ exhibits the desired target attribute values $\{ t_k \}_{k=0}^{K-1}$ while preserving the identity represented by the original iris image $x_0$ (see Fig. \ref{fig:teaser}). We formulate this as the minimization of a composite loss function:
\begin{align}
\mathcal{L}(w) \;=\;&
\underbrace{\sum_{k=0}^{K-1}
\mathcal{L}_{\mathrm{attr}:k}\big(a_k(G(w)),\, t_k \big)}_{\text{attribute losses}}
\nonumber \\[4pt]
&+ 
\underbrace{
\mathcal{L}_{\mathrm{id}}\big(
\phi_{\mathrm{id}}(G(w)),\, \phi_{\mathrm{id}}(x_0) \big)
}_{\text{identity preservation loss}}
\;
+\;
\underbrace{
\tfrac{1}{2}\,\lVert w - w_0 \rVert^2
}_{\text{latent regularization}},
\label{eqn:composite}
\end{align}

\noindent
where:
\begin{itemize}
    \item $\mathcal{L}_{\mathrm{attr}: k}$ is a differentiable loss between the current and target value of attribute $k$ calculated for the current and target synthetic iris images,
    \item $\mathcal{L}_{\mathrm{id}}$ enforces the identity preservation and is the distance between two iris images, $x=G(w)$ and $x_0$, and 
    \item $0.5\lVert w - w_0 \rVert^2$ is the latent regularization term to constrain the update steps in the latent space and thus ensure that $w^\star$ does not deviate excessively from $w_0$.
\end{itemize}

\noindent
The optimization problem is then given by:
\[
    w^\star \;=\; \arg\min_{w \in \mathcal{W}} \; \mathcal{L}(w) .
\]

\subsection{Definitions of Single-Attribute Loss Terms}
\label{sec:Formula}

\paragraph{Identity Preservation Loss:} 

Identity preservation is achieved by applying a set of iris recognition-specific filters $\mathcal{F}_\mathrm{iris}$ to extract iris features $\phi_{\mathrm{id}}$, namely:

\begin{displaymath}
    \phi_{\mathrm{id}}(x) = \mathcal{F}_\mathrm{iris} \circledast \psi(x),
\end{displaymath}

\noindent
where $\psi(x)$ denotes Daugman's iris normalization function \cite{daugman1993high}, which maps the $640\times480$ iris image expressed in Cartesian to a fixed-size $512\times64$ representation in polar coordinate system, $\mathcal{F}_\mathrm{iris}$ is composed of Gabor wavelets from the OSIRIS~\cite{othman2016osiris} and human perception-driven  \cite{czajka2019domain} matchers, and $\circledast$ denotes the convolution operation. The identity preservation loss is then defined as:
\begin{equation}
    \label{eq:id_Loss}
    \mathcal{L}_{\mathrm{id}}(x,x_0) = \|\,\phi_{\mathrm{id}}(x) - \phi_{\mathrm{id}}(x_0)\,\|,
\end{equation}

\noindent
where $x=G(w)$ is the current synthesized iris image, $x_0$ is the initial iris image, and $\|\cdot\|$ denotes the $L_1$ norm.

\paragraph{Mask Loss:} There may be a need to preserve the iris shape when manipulating other attributes. This is achieved by introducing a binary cross-entropy loss between the predicted mask logits of the current iris image $x$ and the mask of the initial iris image $x_0$, namely:

\begin{align} 
    \label{eq:maskLoss}
    \mathcal{L}_{\mathrm{attr}:\mathrm{mask}} \;=\;& \; -\Big[\mu(x_0)\log\big(\tilde\mu(x)\big)\\ \nonumber
    & +\;(1-\mu(x_0))\log\big(1-\tilde\mu(x)\big)\Big],
\end{align}

\noindent
where $\tilde\mu(x)\in[0.0,1.0]$ is the map providing the probability of every pixel representing the iris texture, where $\tilde\mu(x) = 1.0$ represents iris texture, and $\tilde\mu(x) = 0.0$ represents non-iris pixels, and $\mu(x_0)$ is the ground-truth binary iris mask for image $x_0$.

\paragraph{Sharpness Loss:} To adjust the sharpness of the iris image, we adopt the iris image sharpness definition provided by the ISO/IEC 19794-6 iris quality standard \cite{iso2011iris}. According to this definition, instead of simply measuring the signal's energy located within the highest spatial frequencies, sharpness quantifies the power spectrum within the selected frequency band (defined by a single kernel $\mathcal{F}_\mathrm{sharpness}$ in \cite{iso2011iris}), within which the iris identity features are usually extracted. Following \cite{iso2011iris}, we define the iris image sharpness loss as:

\begin{equation}
        \mathcal{L}_{\mathrm{attr}:\mathrm{sharpness}} = \big\| 100 \cdot P^2/(P^2 + C^2) - s^* \big\|,
\end{equation}

\noindent
where:

\begin{itemize}
    \item $\|\cdot\|$ denotes the $L_1$ norm,
    \item $s^*$ denotes the target sharpness value,
    \item $C$ = 1,800,000 as recommended in \cite{iso2011iris},
    \item $P=S/|\mu(x)|$ is the signal's power,
    \item $\mu(x)$ is the binary mask of iris image $x$, and $|\cdot|$ denotes the number of pixels corresponding to the iris texture,
    \item $S$ is a squared sum of the elements within the masked filtered iris image after convolving the image $x$ with the sharpness kernel $\mathcal{F}_\mathrm{sharpness}$:

    \begin{displaymath}
        S = \sum_{i=0}^{W-1}\sum_{j=0}^{H-1} \big[(x \circledast \mathcal{F}_\mathrm{sharpness})(i,j)\big]^2.
    \end{displaymath}
    
\end{itemize}

Thus, our adaptation of ISO sharpness metric (a) makes it differentiable with respect to the generative model latent space (by removing the rounding to the closest integer value, as recommended in \cite{iso2011iris}), and (b) allows for assessing sharpness only within the iris texture area by incorporating the binary mask $\mu(x)$ and normalizing $S$ by the number of pixels within this iris mask, instead of normalizing by image height and width, as originally recommended by the ISO standard \cite{iso2011iris}.


\paragraph{Pupil and Iris Size Loss:} The pupil size is defined as the radius of the circle that approximates the pupil-iris boundary, while the iris size is the radius of a circle that approximates the iris-sclera boundary \cite{iso2011iris}. We use a differentiable U-Net-based regression model from the open-source iris recognition package \cite{ND_OpenSourceIrisRecognition_GitHub} estimating circular approximations of iris and pupil circles and their radii, $r_\mathrm{pupil}(x)$ and $r_\mathrm{iris}(x)$, for the iris image $x$. The appropriate loss terms are defined as:

\begin{equation}
     \mathcal{L}_{\mathrm{attr}:\mathrm{pupil}} = \|\,r_\mathrm{pupil}(x) - r_\mathrm{pupil}^*\,\|,
\end{equation}
\noindent
\begin{equation}
     \mathcal{L}_{\mathrm{attr}:\mathrm{iris}} = \|\,r_\mathrm{iris}(x) - r_\mathrm{iris}^*\,\|,
\end{equation}

\noindent
where $r_\mathrm{pupil}^*$ and $r_\mathrm{iris}^*$ denote target values of pupil or iris, respectively.

\paragraph{Pupil-to-Iris Ratio Loss:}

It is sometimes more convenient or required to control the pupil-to-iris dilation ratio, instead of pupil size and iris size independently. Having the $r_\mathrm{pupil}(x)$ and $r_\mathrm{iris}(x)$ estimated, and following \cite{iso2011iris}, we define the pupil-to-iris ratio loss
term as:
\begin{equation}
    \mathcal{L}_{\mathrm{attr}:\mathrm{PIR}} = \bigg\|\,100\cdot\frac{r_\mathrm{pupil}}{r_\mathrm{iris}+\epsilon} - \mathrm{t}\,\bigg\|,
\end{equation}

\noindent
where $t$ denotes the target value of pupil-to-iris ratio and $\epsilon = 10^{-6}$ prevents division by zero.

\subsection{Selected Combinations of Loss Terms}

As indicated previously, the proposed gradient-guided image attribute manipulation does not assume a strong disentanglement of representations of the attributes in the latent space. Thus, the control of one attribute may inadvertently impact another one. To achieve the desired single-attribute control, one may need to compose several loss functions counteracting not desired changes of attributes. 

One example is the control of sharpness, for which we observed simultaneous and not desired changes in iris shape during the latent space traversal. Combining $\mathcal{L}_{\mathrm{attr}:\mathrm{sharpness}}$ with $\mathcal{L}_{\mathrm{{attr}:{mask}}}$, which penalizes for deviations from the initial iris shape, allows the model to find appropriate path in the latent space to change the iris texture sharpness only.

\subsection{Optimization Procedure}
To generate synthetic iris samples, we employed AdamW optimizer with a learning rate of $0.008$. The optimization process was performed for a maximum of 30 iterations. However, a target criterion can be specified to enable early termination once the desired objective values of the selected set of image attributes are reached. To enhance training stability and mitigate the risk of gradient explosion, gradient clipping was applied to the latent variable, with the maximum norm constrained to $1.0$.

\section{Results}
\label{sec:Results}

\subsection{Visualization of Image Attribute Manipulation} 

\begin{figure*}[!htbp]
    \begin{center}
    \includegraphics[width=0.95\textwidth]{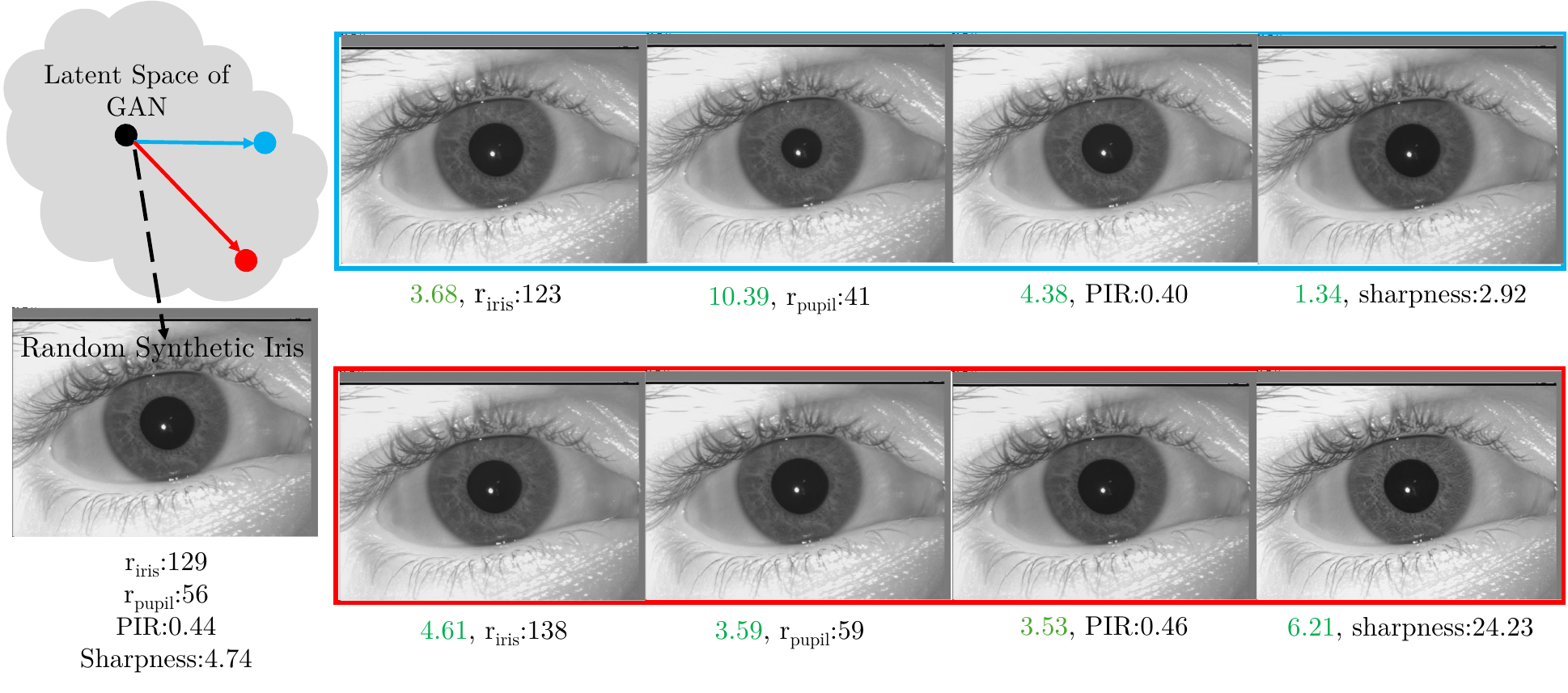}
    \caption{Illustration of the iris image attribute manipulation process through gradient-guided traversal through the $\mathcal{W}$ latent space of the StyleGAN2-ADA model trained for iris image synthesis, highlighting the effect of incorporating an identity loss term. The first row, bordered in blue, illustrates the resulting images when a decrease in the relevant attribute was requested. The second row, bordered in red, illustrates images with these attributes increased. The numerical values under each image indicate the Euclidean distance obtained with the {\it TripletNN} matcher, and the corresponding attribute value obtained after latent space traversal.}
    \label{fig:iris_changes}
\end{center}
\end{figure*}

To conduct experiments, we first trained a StyleGAN2-ADA generative model \cite{Karras_neurips_2020} on a combined set of publicly-available iris images. Figure~\ref{fig:iris_changes} demonstrates the image attribution manipulation process for selected properties of the synthesized iris image. The values associated with each image represent the Euclidean distance between embeddings of the original and manipulated iris images (the lower the distance, the better the match, and the green number indicates that the original iris identity has been successfully preserved), as well as the target values of iris attributes obtained after optimization. 

To assess how the identity is preserved, in all evaluations we applied the {\it TripletNN} matcher \cite{ND_OpenSourceIrisRecognition_GitHub}. This has been included into the NIST's IREX X leaderboard \cite{irexx}, thus represents a matcher tested on a large number of samples in a well-regarded iris recognition evaluation program, and it was not used in the method design, \eg in definition of identity loss $\mathcal{L}_{\mathrm{{attr}:{id}}}$. 

As can be seen, using the proposed gradient-guided latent space traversal, it is possible to change selected geometrical (iris radius, pupil radius, and pupil-to-iris ratio) and textural (iris texture sharpness) attributes, at the same time preserving the identity-related features.

\subsection{Identity Preservation}

Since the biometric utility of the manipulated samples is higher if the identity-related features of the initial iris image are preserved, the results of the proposed method are analyzed in two scenarios: with and without the use of the identity loss term $\mathcal{L}_{\mathrm{{attr}:{id}}}$. This was done to investigate the impact of identity preservation on the latent space traversal, and how accurately the proposed mechanism can preserve identity features during the generation of a new iris image with manipulated attributes.

\begin{figure}[!htbp]
    \begin{center}
    \includegraphics[width=\linewidth]{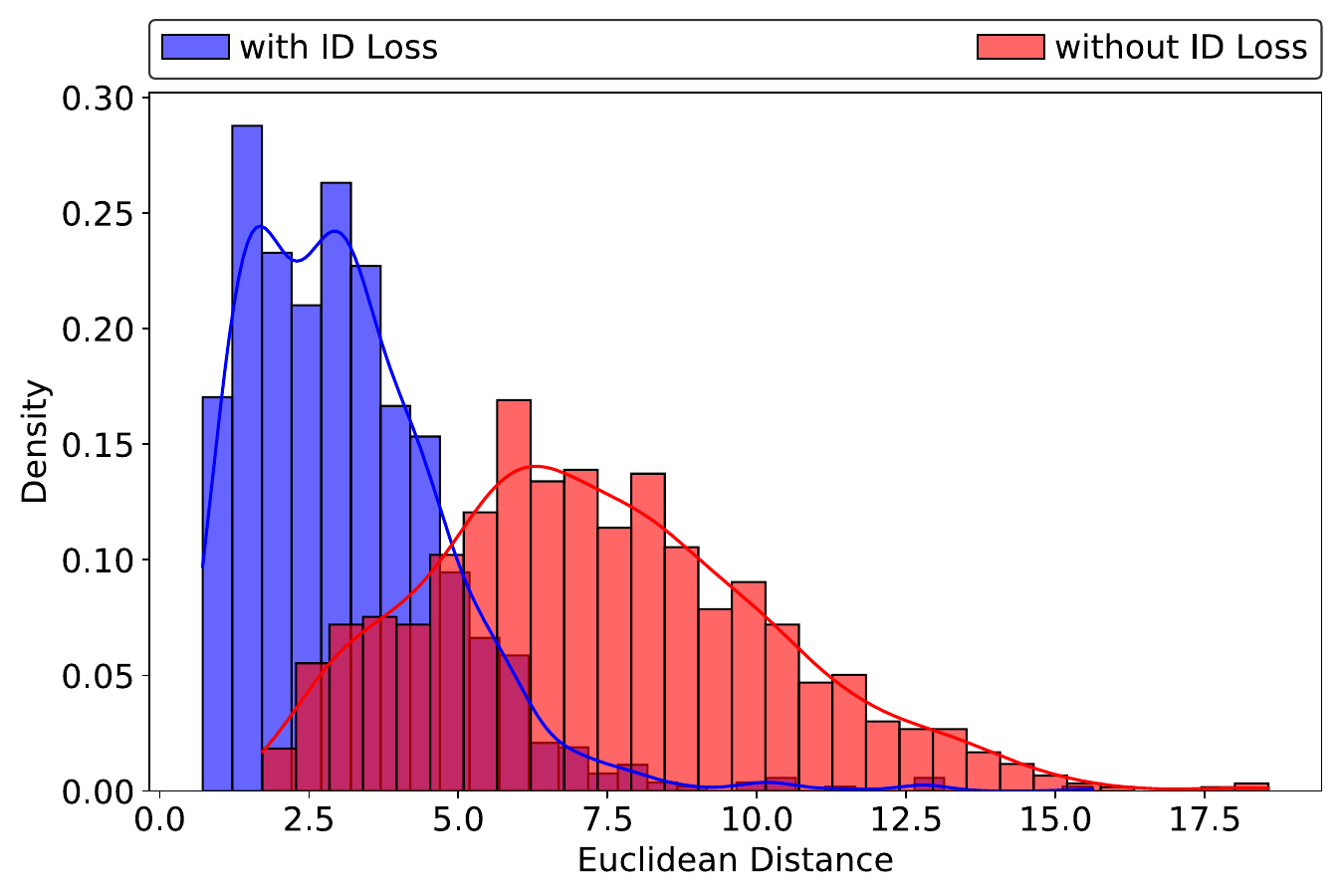}    
    \caption{Euclidean distance distributions between the embeddings of the initial iris image and the final image with selected attributes manipulated, obtained with {\it TripletNN} matcher \cite{ND_OpenSourceIrisRecognition_GitHub} in two scenarios: with and without inclusion of identity loss component $\mathcal{L}_{\mathrm{{attr}:{id}}}$. Plots were obtained for 1,200 comparison scores in each scenario: 150 random latent codes $\times$ 4 manipulated attributes (pupil size, iris size, sharpness and pupil-to-iris ratio) $\times$ 2 (decreasing and increasing the attribute's value). For the {\it TripletNN} matcher, Euclidean distances above 12 indicate a non-match, thus a failure to preserve identity.}  
    \label{fig:euclidean_distance}
\end{center}
\end{figure}

To do this, we started with $150$ random latent codes (initial positions $w_0$ in the latent space) for each attribute modification experiment. The final sample pairs were selected to have nearly identical attribute values, ensuring a fair comparison. The Euclidean distance between the embeddings of the initial iris image $x_0$ and the final manipulated image $x^\star$ was calculated by using the {\it TripletNN} matcher, and distributions of the distances for scenarios with and without use of identity loss are illustrated in Fig.~\ref{fig:euclidean_distance}. The blue and red bars (with corresponding PDF approximations) denote samples synthesized with and without an identity loss component, respectively. The distributions reveal a clear shift. Specifically, samples generated with identity loss exhibit lower distance, which is indicative of a higher degree of identity preservation compared to the initial iris image. The distances for samples generated without identity loss are skewed toward higher values, reflecting a reduced level of identity preservation.

\begin{figure*}[!htbp]
    \begin{center}
    \includegraphics[width=\linewidth]{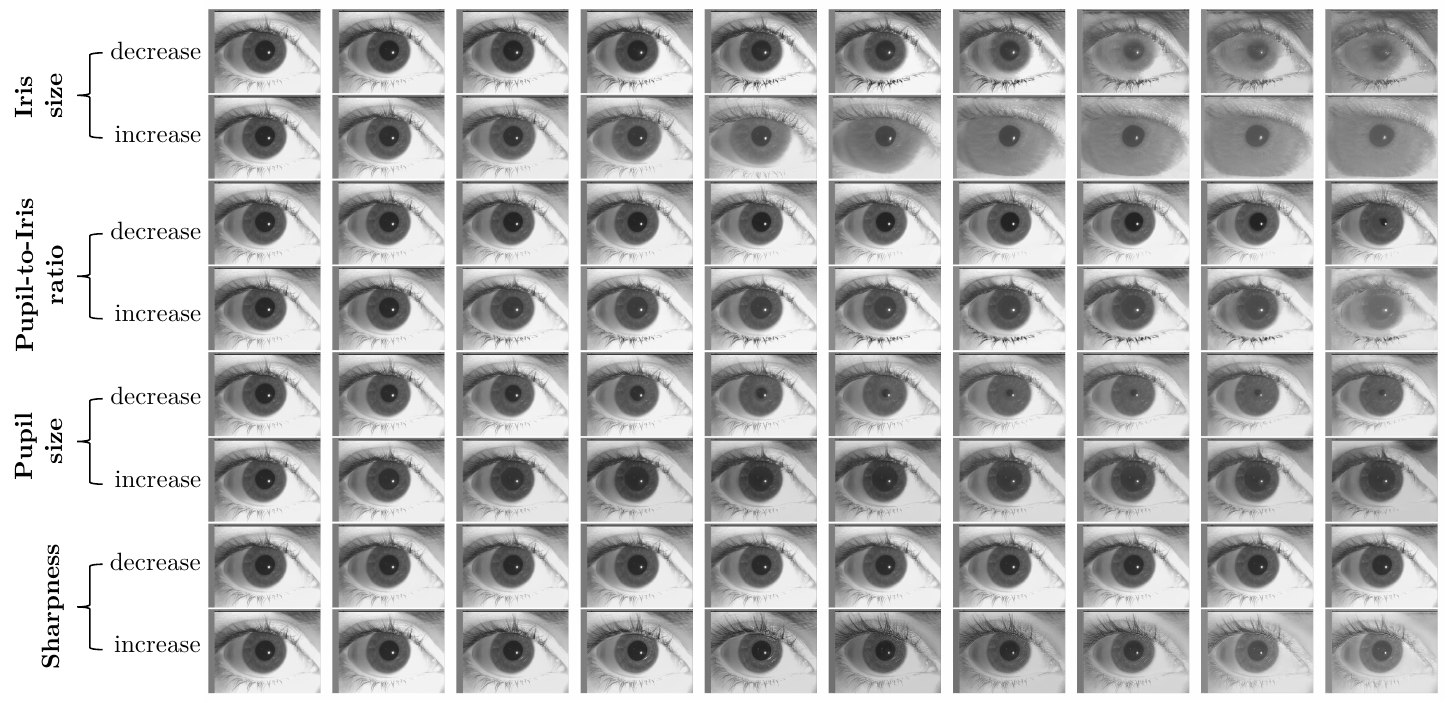}
    \caption{\label{fig:id_noid_samples} Illustration of extensive (for 600 optimization steps in this Figure) manipulations of synthetic iris image attributes. This may result in large deviations from the initial latent code, what in consequence may degrade the original iris structure and moves the optimization toward regions of the latent space $\mathcal{W}$ that do not produce valid iris samples.}
\end{center}
\end{figure*}

Incorporating identity loss constrains latent space traversal, preventing the generation of images that significantly diverge from the original identity unless the latent code moves too far from its initial position. Fig.~\ref{fig:id_noid_samples} illustrates generated samples with the amplitude of manipulated attributes being exaggerated, showing that significant changes in iris attributes fail to produce images closely resembling the original sample.

\subsection{``Z'' or ``W'' StyleGAN Latent Space: Which One Is Better?}

StyleGAN2-ADA, used in this study as an example generative model, employs two latent spaces: the input latent space $\mathcal{Z}$ and the intermediate latent space $\mathcal{W}$, where $\mathcal{Z}$ is transformed into $\mathcal{W}$ via the ``mapping network''. It is claimed that $\mathcal{W}$ offers an improved disentanglement between coarse and fine-grained image features \cite{karras2021alias}. In this work, we primarily focus on $\mathcal{W}$. However, we also compared results when traversing both latent spaces and observed no significant differences in the quality and the identity preservation of the generated samples. The main observation, however, was that traversing $\mathcal{W}$ allowed for faster convergence towards the desired target attribute.

This confirms the hypothesis that gradient-based traversal of latent spaces is relatively robust against the structure of the latent space, and thus bodes well for generalization of this approach to a wide spectrum of generative decoders.

\begin{figure*}[!htbp]
    \begin{center}
    \includegraphics[width=\linewidth]{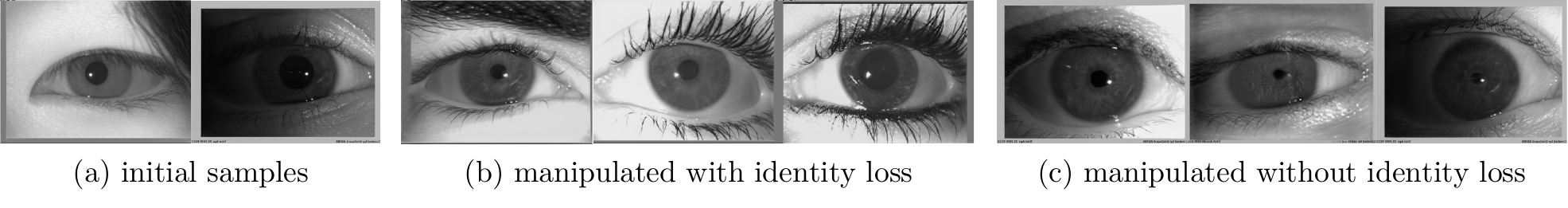}
    \caption{\label{fig:low_quality}Samples of synthetic iris images with overall quality score less than 10. (a) iris images from initial latent codes, (b) samples from $\mathcal{W}$-space with identity loss, (c) samples from $\mathcal{W}$-space without identity loss.} 
\end{center}
\end{figure*}

\subsection{Quality Assessment for Synthetic Iris Images}
\label{quality_assessment}

Popular general-purpose image quality metrics, such as Fréchet Inception Distance (FID), are not well-suited for assessing the utility of biometric samples. These metrics, designed primarily for natural images and relying on feature extractors pretrained on datasets like ImageNet, fail to capture the near-infrared iris image attributes that are critical for iris recognition. Instead, to evaluate the quality of the manipulated samples, we use the ISO/IEC 19794-6 metrics, which provide domain-specific quality criteria designed for iris recognition.

We optimized 150 random latent codes for 30 steps and computed the quality scores of the resulting images. The percentage of samples with the ISO/IEC 19794-6 overall quality score below 10 was 1.5\% when traversing the $\mathcal{W}$ space with identity loss, 2.7\% for traversing $\mathcal{W}$ without identity loss, and 1.4\% for traversing the $\mathcal{Z}$ space with identity loss. Inspection of samples with quality scores less than 10 showed that some of these originated from low-quality images corresponding to initial latent codes (Fig.~\ref{fig:low_quality}a), while the remaining samples were generated during optimization (Fig.~\ref{fig:low_quality}b–c). The percentage of manipulated samples with ISO/IEC 19794-6 overall quality scores above 70 were 98\%, 95\%, and 98\%, respectively. These results highlight very high chances of obtaining good-quality iris images after the attribute manipulation process is completed. 

\section{Discussion and Conclusions}

This work introduces the first known to us iris image augmentation strategy that leverages a gradient-guided exploration of the latent space of pre-trained generative image decoders, such as those included in generative adversarial networks (GAN). Our approach allows us to traverse the decoder's latent space in specific directions, guided by the gradients of the selected iris image attribute with respect to the model's latent space. Additionally, the proposed method adjusts the attribute's value toward a desired target while preserving the original iris identity, which allows to control attributes of both same-eye and different-eye samples. We validated this method using attributes such as pupil size, iris size, pupil-to-iris ratio, and image sharpness. However, this technique isn't limited to these attributes, or to the iris recognition domain, and it may be applied to any image and attribute that can be formulated as a function differentiable with respect to the decoder's latent space.

One application of this approach is its ability to manipulate both synthetic and real-world iris images. Rather than relying solely on random latent codes to generate iris images, inversion techniques can be employed to project any existing iris image into the latent space and extract its corresponding latent code. This latent code can then be fed back into a pre-trained generative model, enabling precise modification of the synthesized image’s attributes.

Unlike facial attributes (\eg expression, hair, glasses, and pose, which are easily perceived), subtle changes to the iris, such as variations in sharpness of the iris texture, are not often apparent to the naked eye and require magnification. The limited number of easily definable, semantically meaningful, and perceptible iris features makes it difficult to create a robust attribute-based latent space. However, the current solution is designed to specifically mitigate these challenges and guide the latent space traversal towards representations of images having the desired attributes. 

By generating iris images controlled by specific ISO-defined quality metrics, the proposed method aids in enhancing the diversity of biometric datasets. This, in turn, can potentially improve the robustness of fully data-driven biometric recognition approaches and serve as a mechanism to synthesize privacy-safe samples used in human examiners training or educational materials (\eg IREX V poster \cite{irexv}, which includes real iris images and is used in instructional materials within law enforcement agencies implementing iris recognition). 

Related to the above and a persistent {\bf limitation} of GAN-sourced decoders, used in this work to validate the approach, is that not all random latent codes generate high-quality images due to inherent randomness (\eg 2 of the 150 generated codes resulted in low quality synthesized images in our experiments). Therefore, initial latent codes should be selected from those that generated images that pass basic quality checks, which depend on the application domain, and in case of iris image synthesis these may be a subset of ISO/IEC 29794-6 metrics. 

We offer the source codes of the proposed method, and a dataset of 4,500 synthesized iris images used in Sec. \ref{quality_assessment} for replicability purposes, and to facilitate further research. 

In future work, we plan to use these modified samples to train iris presentation attack detection systems to assess how the proposed pipeline enhances detection performance and robustness, while contributing to the development of privacy-preserving frameworks such as those introduced in~\cite{mitcheff2024privacy}.

{\small
\bibliographystyle{ieee}
\bibliography{egbib}
}

\end{document}